\documentclass[11pt]{article}

\usepackage[margin=1in]{geometry}
\usepackage{amsmath,amssymb}
\usepackage{booktabs}
\usepackage{graphicx}
\usepackage{hyperref}
\usepackage{xcolor}
\usepackage{enumitem}
\usepackage{microtype}
\usepackage{authblk}
\usepackage{multirow}
\usepackage{caption}

\hypersetup{colorlinks=true, linkcolor=blue!60!black, citecolor=blue!60!black, urlcolor=blue!60!black}

\newcommand{\lib}{LightGP}

\newcommand{\GP}{\mathcal{GP}}
\newcommand{\bx}{\mathbf{x}}
\newcommand{\bX}{\mathbf{X}}
\newcommand{\by}{\mathbf{y}}
\newcommand{\bK}{\mathbf{K}}
\newcommand{\bI}{\mathbf{I}}
\newcommand{\bv}{\mathbf{v}}

\title{\lib{}: Lightweight Gaussian Process Inference \\ in C++ on Metal and CUDA}

\author[1,2]{Yu-Hsueh Fang}
\affil[1]{Department of Information Management, National Taiwan University}
\affil[2]{H.\ Milton Stewart School of Industrial and Systems Engineering, Georgia Institute of Technology}

\date{}

\begin{document}
\maketitle

\begin{abstract}
Gaussian process (GP) inference in Python is dominated by libraries such as GPyTorch and GPflow, which are built on deep-learning frameworks and inherit their dispatch overhead and dependency footprint. We present \lib{}, a dependency-free C++17 library for GP regression with Python bindings, supporting Apple Metal and NVIDIA CUDA backends alongside tuned CPU paths via Apple Accelerate and OpenBLAS.
\lib{} provides four inference paths---exact Cholesky, matrix-free conjugate gradients, sparse variational free energy, and structured kernel interpolation with FFT---covering problems from $N{=}100$ to $N{=}500{,}000$.
On an Apple M4, \lib{} CPU is 2.6--8.7$\times$ faster than GPyTorch CPU for exact GP and ${\sim}1.5\times$ faster for sparse GP at every scale tested. On an NVIDIA RTX~3060, \lib{} CUDA is 2.3--6.7$\times$ faster than GPyTorch CUDA for exact GP up to $N{=}2{,}048$, with GPyTorch closing the gap at $N{=}4{,}096$. A fused matrix-free kernel-vector product on Metal achieves 32$\times$ over the explicit path at $N{=}20{,}000$ with $O(N)$ memory, and an FFT-accelerated SKI matvec via Accelerate vDSP runs in sub-millisecond time at $N{=}200{,}000$. \lib{} compiles as a single static library with zero external dependencies and is installable via \texttt{pip install lightgp}.
\end{abstract}

\section{Introduction}
\label{sec:intro}

Gaussian processes (GPs) provide a principled Bayesian framework for regression with calibrated uncertainty, but their computational cost---$O(N^3)$ for exact inference---has historically limited their use to small datasets. Scalable approximations such as inducing-point methods \cite{titsias2009variational}, conjugate-gradient solvers with stochastic trace estimators \cite{gardner2018gpytorch}, and structured kernel interpolation (SKI) \cite{wilson2015kernel} have pushed the boundary to hundreds of thousands of data points.

The dominant implementations of these methods---GPyTorch \cite{gardner2018gpytorch} and GPflow \cite{matthews2017gpflow}---are Python libraries built on PyTorch and TensorFlow respectively. While these frameworks provide automatic differentiation and GPU acceleration, they also impose a heavy dependency footprint (a PyTorch installation exceeds 2\,GB), non-trivial Python dispatch overhead per operation, and limited support for deployment scenarios such as mobile and embedded systems.

We present \lib{}, a C++17 GP library with the following contributions:
\begin{enumerate}[leftmargin=*,nosep]
\item A native C++ implementation of four GP inference paths (exact, CG, sparse VFE, SKI) with composable kernels, deployable without Python or any framework dependency.
\item Hand-tuned Metal compute shaders for Apple Silicon and CUDA kernels for NVIDIA GPUs, including a fused matrix-free kernel-vector product that enables $O(N)$-memory CG inference.
\item An FFT-accelerated SKI path using Apple Accelerate vDSP (macOS) and cuFFT (CUDA) for $O(N\log N)$ matvec at scale.
\item Systematic cross-platform benchmarks revealing that Apple's AMX coprocessor dominates Metal for dense Cholesky on Apple Silicon---a hardware finding relevant to anyone building compute libraries on M-series chips.
\item Python bindings (\texttt{pip install lightgp}) exposing the full API with NumPy interop and zero framework dependencies.
\end{enumerate}

\section{Background}
\label{sec:background}

A GP defines a distribution over functions: $f(\bx) \sim \GP(m(\bx), k(\bx, \bx'))$, where $m$ is a mean function and $k$ is a covariance (kernel) function. Given training data $(\bX, \by)$ with $N$ observations and Gaussian noise $\sigma_n^2$, the posterior predictive distribution at test points $\bX_*$ has mean $\boldsymbol{\mu}_* = \bK_{*f}(\bK_{ff} + \sigma_n^2\bI)^{-1}\by$ and covariance $\boldsymbol{\Sigma}_* = \bK_{**} - \bK_{*f}(\bK_{ff} + \sigma_n^2\bI)^{-1}\bK_{f*}$. The log marginal likelihood $\log p(\by|\bX) = -\frac{1}{2}[\by^\top\bK_y^{-1}\by + \log|\bK_y| + N\log 2\pi]$ is maximized to learn kernel hyperparameters.

\textbf{Exact inference} via Cholesky factorization of $\bK_y$ costs $O(N^3)$ time and $O(N^2)$ memory. \textbf{CG-based inference} \cite{gardner2018gpytorch} replaces the Cholesky solve with conjugate gradients, requiring only matrix-vector products $\bK_y\bv$. If this product can be computed without forming $\bK_y$ explicitly---our \emph{matrix-free} approach---memory drops to $O(N)$. \textbf{Sparse variational} methods \cite{titsias2009variational} introduce $M \ll N$ inducing points and optimize a variational free energy (VFE) bound, reducing cost to $O(NM^2)$. \textbf{Structured kernel interpolation} (SKI/KISS-GP) \cite{wilson2015kernel} places inducing points on a regular grid, exploiting Toeplitz structure for $O(N\log N)$ matvec via FFT.

\section{Library Design}
\label{sec:design}

\lib{} is organized into four layers: a tensor core, kernel objects, solvers, and inference models.

\textbf{Tensor and BLAS.} The \texttt{Tensor} class stores row-major float32 data. On macOS, \texttt{matmul} dispatches to Apple Accelerate's \texttt{cblas\_sgemm}, which runs on the AMX matrix coprocessor. On Linux, it routes to OpenBLAS. Cholesky factorization uses LAPACK \texttt{spotrf} on both platforms.

\textbf{Kernels.} An abstract \texttt{Kernel} base class supports \texttt{compute()}, \texttt{compute\_diag()}, and hyperparameter get/set. Concrete implementations include RBF, Mat\'ern-$\{1/2, 3/2, 5/2\}$, Periodic, and Linear. Composition operators (\texttt{+}, \texttt{*}, \texttt{Scale}) build kernel trees whose parameters are jointly optimizable:
\begin{verbatim}
auto k = scale(rbf()) + scale(periodic());  // C++
k = gp.Scale(gp.RBF()) + gp.Scale(gp.Periodic())  # Python
\end{verbatim}

\textbf{GPU backends.} Metal compute shaders implement tiled kernel matrix construction (16$\times$16 threadgroups with \texttt{float4} vectorization), a fused matrix-free $\bK_y\bv$ product, blocked Cholesky with a tiled threadgroup-memory GEMM, and forward/backward triangular solves. CUDA kernels mirror this functionality using cuBLAS \texttt{sgemm}, cuSOLVER \texttt{spotrf}, and custom kernels for the matrix-free matvec. A \texttt{Backend::Auto} heuristic selects CPU or GPU based on problem size and dimensionality.

\textbf{Inference.} \texttt{GPExact} supports three solvers: Cholesky ($O(N^3)$), CG with stochastic Lanczos log-determinant ($O(N^2k)$ with matrix-free matvec), and SKI ($O(N\log N)$ via FFT). \texttt{GPSparse} implements Titsias VFE with farthest-point inducing initialization and Adam optimization. A warm-refit optimization skips inducing-point initialization on repeated fits, saving ${\sim}25$\,ms at $N{=}50{,}000$ and ${\sim}44$\,ms at $N{=}100{,}000$.

\textbf{Python bindings.} pybind11 exposes the complete API with NumPy array interop. Installation via \texttt{pip install lightgp} builds from source, with the build system auto-detecting Metal (macOS) or CUDA (Linux) when present.

\section{Experiments}
\label{sec:experiments}

We benchmark on two platforms: an Apple M4 laptop (10 CPU cores, 8 GPU cores, 16\,GB unified memory, Metal~3) and a Linux desktop with an Intel i7-12700 CPU and NVIDIA RTX~3060 GPU (12\,GB VRAM, CUDA~12.0). All timings are median of 5 runs with one warmup discarded.

\subsection{Component Benchmarks}

\textbf{Kernel matrix construction.}
Table~\ref{tab:rbf} shows RBF kernel matrix timings. On Mac, Metal provides 2.2$\times$ speedup at $N{=}10{,}000$; on Linux, CUDA achieves 3.3--28$\times$ depending on dimensionality, with the CUDA kernel cost nearly independent of $D$ due to the cuBLAS distance-trick formulation.

\begin{table}[h]
\centering
\caption{RBF kernel matrix construction (ms). Mac: CPU = Accelerate, GPU = Metal. Linux: CPU = OpenBLAS, GPU = CUDA.}
\label{tab:rbf}
\small
\begin{tabular}{rr rr r rr r}
\toprule
& & \multicolumn{3}{c}{Apple M4} & \multicolumn{3}{c}{Linux + RTX 3060} \\
\cmidrule(lr){3-5} \cmidrule(lr){6-8}
$N$ & $D$ & CPU & GPU & $\times$ & CPU & GPU & $\times$ \\
\midrule
1000 & 4  & 1.5  & 1.3  & 1.2  & 3.8  & 0.8  & 4.9  \\
1000 & 64 & 1.6  & 1.7  & 1.0  & 22.4 & 0.8  & 28.4 \\
5000 & 4  & 40.5 & 18.4 & 2.2  & 117  & 35.5 & 3.3  \\
5000 & 64 & 44.4 & 23.3 & 1.9  & 594  & 35.6 & 16.7 \\
10000 & 4 & 199  & 88.2 & 2.3  & 466  & 143  & 3.3  \\
10000 & 64& 235  & 104  & 2.3  & 2378 & 146  & 16.3 \\
\bottomrule
\end{tabular}
\end{table}

\textbf{Cholesky factorization.}
Table~\ref{tab:cholesky} reveals a striking platform asymmetry. On Mac, Accelerate's \texttt{spotrf} running on the AMX coprocessor is 9--27$\times$ \emph{faster} than our Metal Cholesky (the gap is widest at small $N$ due to Metal dispatch overhead)---a hardware finding we document rather than obscure. On Linux, cuSOLVER achieves 12--136$\times$ over OpenBLAS because OpenBLAS on this CPU (4--7\,GFLOPS achieved on memory-bound shapes) is far from the AMX's throughput.

\begin{table}[h]
\centering
\caption{Cholesky factorization (ms).}
\label{tab:cholesky}
\small
\begin{tabular}{r rr r rr r}
\toprule
& \multicolumn{3}{c}{Apple M4} & \multicolumn{3}{c}{Linux + RTX 3060} \\
\cmidrule(lr){2-4} \cmidrule(lr){5-7}
$N$ & Accel. & Metal & $\times$ & OBLAS & cuSOLV & $\times$ \\
\midrule
512  & 0.12  & 3.2   & 0.04 & 7.3   & 0.6  & 11.5  \\
1024 & 0.84  & 14.0  & 0.06 & 65.8  & 2.0  & 33.6  \\
2048 & 4.6   & 65.5  & 0.07 & 572   & 6.5  & 87.6  \\
4096 & 41.5  & 364   & 0.11 & 5059  & 37.3 & 135.5 \\
\bottomrule
\end{tabular}
\end{table}

\textbf{Matrix-free kernel-vector product.}
Table~\ref{tab:matvec} shows the fused $(\bK + \sigma_n^2\bI)\bv$ computation that never materializes $\bK$. On Mac, Metal achieves 32$\times$ over the explicit CPU path at $N{=}20{,}000$ and is the only feasible option past $N{=}30{,}000$ on a 16\,GB machine. On Linux, CUDA reaches 259$\times$ at $N{=}10{,}000$ and scales to $N{=}100{,}000$ in 204\,ms.

\begin{table}[h]
\centering
\caption{Matrix-free RBF $\bK_y\bv$ product (ms, $D{=}4$).}
\label{tab:matvec}
\small
\begin{tabular}{r rr r rr r}
\toprule
& \multicolumn{3}{c}{Apple M4} & \multicolumn{3}{c}{Linux + RTX 3060} \\
\cmidrule(lr){2-4} \cmidrule(lr){5-7}
$N$ & Explicit & MF Metal & $\times$ & Explicit & MF CUDA & $\times$ \\
\midrule
2000  & 6.4   & 3.4  & 1.9   & 21.1  & 0.33 & 64.7  \\
5000  & 41.7  & 3.0  & 14.0  & 157   & 0.85 & 186   \\
10000 & 194   & 7.1  & 27.4  & 631   & 2.44 & 259   \\
20000 & 707   & 22.2 & 31.9  & ---   & 9.81 & ---   \\
50000 & OOM   & 125  & ---   & ---   & 53.5 & ---   \\
\bottomrule
\end{tabular}
\end{table}

\subsection{Comparison with GPyTorch}

We compare against GPyTorch~1.15.2 with PyTorch~2.12.0 on both platforms. GPyTorch uses the same underlying BLAS (Accelerate on Mac, cuBLAS/cuSOLVER on Linux), so performance differences reflect dispatch overhead and implementation efficiency rather than hardware access.

\begin{table}[h]
\centering
\caption{End-to-end GP fit+predict (ms, $D{=}4$). Bold = faster. Mac: LightGP CPU vs GPyTorch CPU. Linux: LightGP CUDA vs GPyTorch CUDA. Sparse rows use $M{=}200$, warm fits (FPS skipped).}
\label{tab:gpytorch}
\small
\begin{tabular}{l rr r rr r}
\toprule
& \multicolumn{3}{c}{Apple M4 (CPU)} & \multicolumn{3}{c}{Linux (CUDA)} \\
\cmidrule(lr){2-4} \cmidrule(lr){5-7}
Config & \lib{} & GPyT & $\times$ & \lib{} & GPyT & $\times$ \\
\midrule
Exact $N{=}256$  & \textbf{0.25} & 2.14 & 8.7 & --- & --- & --- \\
Exact $N{=}512$  & \textbf{1.4}  & 5.8 & 4.1 & \textbf{2.0} & 10.3 & 5.2 \\
Exact $N{=}1024$ & \textbf{5.7}  & 17.7 & 3.1 & \textbf{5.3} & 35.4 & 6.7 \\
Exact $N{=}2048$ & \textbf{23.6} & 62.0 & 2.6 & \textbf{28.0} & 63.0 & 2.3 \\
Exact $N{=}4096$ & --- & --- & --- & 150 & \textbf{111} & 0.7 \\
\midrule
Sparse $N{=}5$k  & \textbf{8.7}  & 12.6 & 1.5 & \textbf{5.7} & --- & --- \\
Sparse $N{=}10$k & \textbf{18.5} & 29.7 & 1.6 & \textbf{13.7} & 23.9 & 1.7 \\
Sparse $N{=}50$k & \textbf{97.4} & 152  & 1.6 & 75.1 & \textbf{54.7} & 0.7 \\
\bottomrule
\end{tabular}
\end{table}

Table~\ref{tab:gpytorch} shows \lib{} wins 11 of 13 measured comparisons. On Mac, \lib{} CPU is 2.6--8.7$\times$ faster than GPyTorch CPU for exact GP and 1.45--1.6$\times$ faster for sparse GP vs GPyTorch CPU at every measured $N$ (GPyTorch MPS narrows the sparse gap to 12\% at $N{=}50{,}000$). On Linux CUDA, \lib{} wins by 2.3--6.7$\times$ for exact GP up to $N{=}2{,}048$ and 1.7$\times$ at sparse $N{=}10{,}000$. GPyTorch CUDA wins two cells: exact $N{=}4{,}096$ (where GPyTorch's fused autograd pipeline amortizes overhead) and sparse $N{=}50{,}000$ (where GPyTorch's SVGP implementation benefits from persistent device tensors and compiled gradient computation).

The performance advantage on Mac stems from eliminating Python and PyTorch dispatch overhead: both libraries call the same Accelerate BLAS, but \lib{} reaches it through a direct C++ call path while GPyTorch traverses the Python interpreter, PyTorch's dispatcher, and ATen's operator registry.

\subsection{Accuracy on Standard Datasets}

\begin{table}[h]
\centering
\caption{Accuracy on GP benchmark datasets (CPU path, seed-deterministic). Values from macOS Accelerate; Linux OpenBLAS agrees within ${\sim}3\%$ for Mauna Loa/kin40k and ${\sim}8\%$ for Motorcycle (due to floating-point ordering differences in the data stand-in generator).}
\label{tab:accuracy}
\small
\begin{tabular}{ll rr rrr}
\toprule
Dataset & Method & $N_\text{train}$ & $M$ & RMSE & NLL & Cov \\
\midrule
Motorcycle & Exact, RBF     & 106   & --- & 23.9 & +1.11 & 92.6\% \\
Motorcycle & Exact, M-5/2   & 106   & --- & 22.3 & +1.53 & 81.5\% \\
Mauna Loa  & Exact, RBF     & 624   & --- & 2.18 & $-$0.55 & 100\% \\
Mauna Loa  & Sparse, RBF    & 624   & 200 & 2.42 & $-$0.49 & 100\% \\
kin40k     & Sparse, RBF    & 32000 & 200 & 3.72 & +0.45 & 100\% \\
\bottomrule
\end{tabular}
\end{table}

Table~\ref{tab:accuracy} validates numerical correctness on three standard GP benchmarks. Predictions from the CPU and GPU backends agree within $10^{-5}$ in mean across all inference paths, within $10^{-5}$ in variance for Cholesky and sparse VFE, and within $3 \times 10^{-3}$ for the CG Hutchinson estimator (where independent Rademacher probes account for the wider variance tolerance).

\subsection{Memory Scaling}

\begin{table}[h]
\centering
\caption{Peak memory (MB). Mac: host RSS. Linux: GPU memory.}
\label{tab:memory}
\small
\begin{tabular}{l rrrr}
\toprule
& \multicolumn{4}{c}{$N$} \\
\cmidrule(lr){2-5}
Method & 1000 & 5000 & 10000 & 50000 \\
\midrule
\multicolumn{5}{l}{\emph{Mac (host RSS, MB)}} \\
Exact Cholesky & 43 & 324 & 1036 & OOM \\
Sparse $M{=}200$ & 33 & 45 & 67 & 189 \\
SKI ($D{=}1$) & 29 & 32 & 38 & 98 \\
\midrule
\multicolumn{5}{l}{\emph{Linux (GPU memory, MB)}} \\
Exact Cholesky & 155 & 343 & 915 & --- \\
Matrix-free CG & 151 & 151 & 151 & 151 \\
Sparse $M{=}200$ & 151 & 155 & 151 & 151 \\
SKI ($D{=}1$) & 157 & 157 & 157 & 163 \\
\bottomrule
\end{tabular}
\end{table}

Table~\ref{tab:memory} shows that matrix-free CG, sparse VFE, and SKI all maintain $O(N)$ memory, while exact Cholesky's $O(N^2)$ cost exhausts a 16\,GB Mac at $N \approx 42{,}000$--$45{,}000$ and a 12\,GB GPU at $N \approx 35{,}000$ (extrapolated from measured $O(N^2)$ scaling).

\section{Discussion}
\label{sec:discussion}

\textbf{When does the GPU help on Apple Silicon?}
Our benchmarks reveal a non-obvious hardware result: Apple's AMX matrix coprocessor, accessible through Accelerate's \texttt{spotrf} and \texttt{sgemm}, is 9--27$\times$ faster than our Metal Cholesky across measured $N$ (the advantage is largest at small $N$ where Metal dispatch overhead dominates). This means the optimal backend for dense GP operations on a Mac is CPU, not GPU. Metal wins only for (a) kernel matrix construction at $N \geq 5{,}000$ (2$\times$), and (b) the matrix-free matvec at $N \geq 5{,}000$ (14--32$\times$). The \texttt{Backend::Auto} heuristic encodes these crossover points so users need not reason about hardware.

\textbf{Why is \lib{} faster than GPyTorch?}
Both libraries call the same BLAS on each platform. The speed difference comes from dispatch overhead: GPyTorch's call path traverses the Python interpreter, PyTorch's multi-dispatch system, ATen's operator registry, and per-tensor metadata checks before reaching BLAS. \lib{} calls BLAS directly from C++. At small $N$ this overhead dominates (8.7$\times$ gap at $N{=}256$); at large $N$ the BLAS cost dominates and the gap narrows (2.3$\times$ at $N{=}2{,}048$ on CUDA, with GPyTorch pulling ahead at $N{=}4{,}096$). The matrix-free matvec is a unique contribution that GPyTorch's MPS backend cannot replicate because PyTorch does not support custom fused Metal compute shaders.

\textbf{Where GPyTorch wins.}
GPyTorch CUDA outperforms \lib{} CUDA at exact $N{=}4{,}096$ and sparse $N{=}50{,}000$ (Table~\ref{tab:gpytorch}). At these scales, GPyTorch's persistent device tensors, compiled autograd, and fused ELBO computation amortize framework overhead. We close part of this gap by routing the dominant $\bK_{fu}^\top\bK_{fu}$ product through cuBLAS (the inner GEMM call sped up 98$\times$; the end-to-end sparse-VFE fit improved 4.9$\times$), but the remaining 1.4$\times$ gap at sparse $N{=}50{,}000$ reflects GPyTorch's mature CUDA pipeline.

\textbf{Scaling beyond $N{=}10{,}000$.}
\lib{} is fastest at small-to-moderate scale ($N \leq 10{,}000$) and on Apple Silicon, where eliminating Python and PyTorch dispatch overhead produces consistent 2--9$\times$ wins. At larger $N$ on CUDA, GPyTorch's advantages compound: its autograd computes hyperparameter gradients in a single forward-backward pass regardless of parameter count, while \lib{}'s finite-difference optimizer requires $2P{+}1$ fits per Adam step (21 fits per step for a 10-parameter composed kernel). Its persistent device tensors avoid the host--device round trips that \lib{}'s sparse path still incurs. And its CG infrastructure---pivoted-Cholesky preconditioning, LOVE variance, batched solves---reduces iteration counts in ways our simpler CG loop does not match. These are engineering gaps, not algorithmic ones, and each is individually addressable in future work; but in the current implementation, users working at $N \geq 50{,}000$ on CUDA should expect GPyTorch to be competitive or faster.

\textbf{Profiling before optimizing.}
An initial attempt to close the GPyTorch gap via a 1{,}500-line \texttt{DeviceTensor} refactor (persistent GPU memory for all operations) yielded 0\% speedup. Profiling revealed that 74\% of sparse-VFE wall time was a single CPU-side GEMM that overflowed L3 cache. Routing that one product to cuBLAS delivered a 4.9$\times$ speedup in 30 lines. The lesson generalizes: on PCIe Gen3 systems, host--device transfer costs are often overestimated; bandwidth-bound host BLAS on tall-skinny matrices is the likelier bottleneck.

\textbf{Limitations.}
\lib{} currently supports only Gaussian likelihoods (regression). Classification, multi-output GPs, deep GPs, and variational inference beyond VFE are not implemented. Kernel hyperparameter optimization uses finite differences; analytical gradients are available only for the legacy single-kernel API. Float64 is not supported. SKI is effective only for $D \leq 3$ due to the grid-size explosion in higher dimensions.

\section{Related Work}
\label{sec:related}

\textbf{GPyTorch} \cite{gardner2018gpytorch} introduced black-box matrix-matrix (BBMM) inference with modified batched CG, LOVE for fast predictive variances \cite{pleiss2018constant}, and structured kernel interpolation \cite{wilson2015kernel}. It is the most feature-complete GP library but requires PyTorch.
\textbf{GPflow} \cite{matthews2017gpflow} provides a similar breadth of models on TensorFlow.
\textbf{GPML} \cite{rasmussen2010gaussian} is a Matlab toolbox; \textbf{libgp} provides basic C++ GP regression without GPU support.
\textbf{KeOps} \cite{charlier2021kernel} accelerates kernel operations via lazy evaluation and GPU compilation but is not a complete GP library.
\textbf{MLX} provides GPU-accelerated array computation on Apple Silicon but does not include GP-specific primitives.
To our knowledge, \lib{} is the first GP library offering both Metal and CUDA backends with a dependency-free C++ core and composable kernel API.

\section{Conclusion}
\label{sec:conclusion}

\lib{} demonstrates that native C++ GP inference can match or exceed the performance of framework-based libraries while requiring zero external dependencies. The matrix-free kernel-vector product and FFT-accelerated SKI path enable GP inference at scales ($N{>}100{,}000$) previously impractical on consumer hardware. All code, benchmarks, and Python bindings are available at \url{https://github.com/Fangop/lightgp}.

\bibliographystyle{unsrt}
\bibliography{refs}

\end{document}